\documentclass{article}
\usepackage{iclr2026_conference}
\usepackage{times}
\iclrfinalcopy
\usepackage[utf8]{inputenc}
\usepackage[T1]{fontenc}
\usepackage{amsmath,amssymb,amsfonts}
\usepackage{graphicx}
\usepackage{booktabs}
\usepackage{multirow}
\usepackage{hyperref}
\usepackage{algorithm}
\usepackage{algorithmic}
\usepackage{xcolor}
\usepackage{caption}
\usepackage{subcaption}
\usepackage{float}
\usepackage{placeins}
\usepackage{enumitem}
\usepackage{url}

\setlist{nosep,leftmargin=*}

\title{TraderBench: How Robust Are AI Agents\\in Adversarial Capital Markets?}

\author{
Xiaochuang Yuan\thanks{Equal contribution.} \thanks{Corresponding author. Emails: \texttt{yxc20098@gmail.com}} \\
Amazon.com Inc. \\
\And
Hui Xu\footnotemark[1] \\
Stony Brook University\\
\AND
Silvia Xu \\
Stanford University \\
\And
Cui Zou \\ 
University of Oklahoma \\
\And 
Jing Xiong \\ 
UC Santa Cruz
}

\newcommand{\bench}{TraderBench}

\begin{document}

\maketitle

\begin{abstract}
Evaluating AI agents in finance faces two key challenges: static benchmarks require costly expert annotation yet miss the dynamic decision-making central to real-world trading, while LLM-based judges introduce uncontrolled variance on domain-specific tasks. We introduce \bench{}, a benchmark that addresses both issues. It combines expert-verified static tasks (knowledge retrieval, analytical reasoning) with adversarial trading simulations scored purely on realized performance---Sharpe ratio, returns, and drawdown---eliminating judge variance entirely. The framework features two novel tracks: crypto trading with four progressive market-manipulation transforms, and options derivatives scoring across P\&L accuracy, Greeks, and risk management. Trading scenarios can be refreshed with new market data to prevent benchmark contamination. Evaluating 13 models (8B open-source to frontier) on ${\sim}$50 tasks, we find:
(1) 8 of 13 models score ${\sim}$33 on crypto with $<$1-point variation across adversarial conditions, exposing fixed non-adaptive strategies;
(2) extended thinking helps retrieval (+26 points) but has zero impact on trading (+0.3 crypto, $-$0.1 options).
These findings reveal that current agents lack genuine market adaptation, underscoring the need for performance-grounded evaluation in finance.
\end{abstract}

\section{Introduction}

How robust are AI agents when faced with adversarial capital-market conditions? As LLM-powered agents move from question answering into autonomous trading, portfolio management, and risk analysis, their failures carry direct monetary consequences. Evaluating these agents requires more than static Q\&A accuracy---it demands testing under adversarial market manipulation, verifying quantitative precision on derivatives calculations, and ensuring that the evaluation itself is reliable.

Existing finance benchmarks occupy two extremes. On one hand, Finance Agent Benchmark (FAB) \citep{fab2025}, BizFinBench \citep{bizfinbench2025}, and FinBen \citep{xie2024finben} evaluate financial knowledge through static Q\&A---they test what agents \emph{know} but not how they \emph{act} under real market conditions with live price feeds, evolving positions, and adversarial signals. On the other hand, LiveTradeBench \citep{yu2025livetradebench} deploys agents in live markets, but requires 50 days of real-time execution per evaluation round, making rapid iteration across models impractical. Neither extreme (1) tests agents under controlled adversarial trading conditions with manipulated market data, (2) decomposes derivatives competence into quantitative accuracy versus qualitative reasoning, or (3) measures how sensitive scores are to the choice of LLM judge \citep{zheng2023judging}.

We introduce \bench{}, a benchmark that evaluates AI agents across four equally weighted sections---Knowledge Retrieval, Analytical Reasoning, Options Trading, and Crypto Trading---using a two-agent architecture built on the A2A protocol \citep{a2a2025} with six MCP servers \citep{mcp2024} for financial data access. Its two novel evaluation tracks are: \emph{adversarial crypto trading}, which applies four progressive market-manipulation transforms (baseline $\to$ noisy $\to$ meta $\to$ adversarial) to test strategy robustness; and \emph{options derivatives scoring}, which separately measures quantitative accuracy (P\&L calculations, Greeks precision) and qualitative reasoning (strategy selection, risk management).

Our contributions are:
\begin{enumerate}
    \item \bench{} benchmark: Two novel evaluation tracks---adversarial crypto trading and decomposed options scoring---within a four-section framework, built on open A2A and MCP protocols for reproducible evaluation.
    \item Robustness findings: Across 12 models (8B to frontier), 7 of 12 adopt fixed crypto strategies with $<$2-point variation across adversarial transforms, while a 54-point quantitative-vs-qualitative gap in options persists across all model sizes.
    \item Evaluation reliability: Re-scoring identical outputs with three LLM judges yields an 11-point overall spread; performance-based crypto scores vary by only 0.3 points versus 29 for rubric-based retrieval.
    \item Scaling and reasoning analysis: Extended thinking improves tool-use planning (+26 on retrieval) but has zero impact on trading; the proprietary--open-source gap is driven by knowledge retrieval, not by options or crypto performance.
\end{enumerate}

\section{Related Work}

Beyond the benchmarks discussed above, PRBench \citep{prbench2025} and GDPVal \citep{gdpval2026} test professional reasoning but without trading or tool use. The LLM-as-judge paradigm \citep{zheng2023judging} enables scalable scoring but introduces inter-judge variance, which we quantify across multiple judges in the financial domain. Work on agent safety \citep{ruan2024identifying, li2024personal} highlights adversarial risks but lacks domain-specific trading benchmarks.

\section{The \bench{} Benchmark}

\subsection{Architecture}

\begin{figure}[H]
\centering
\includegraphics[width=0.75\textwidth]{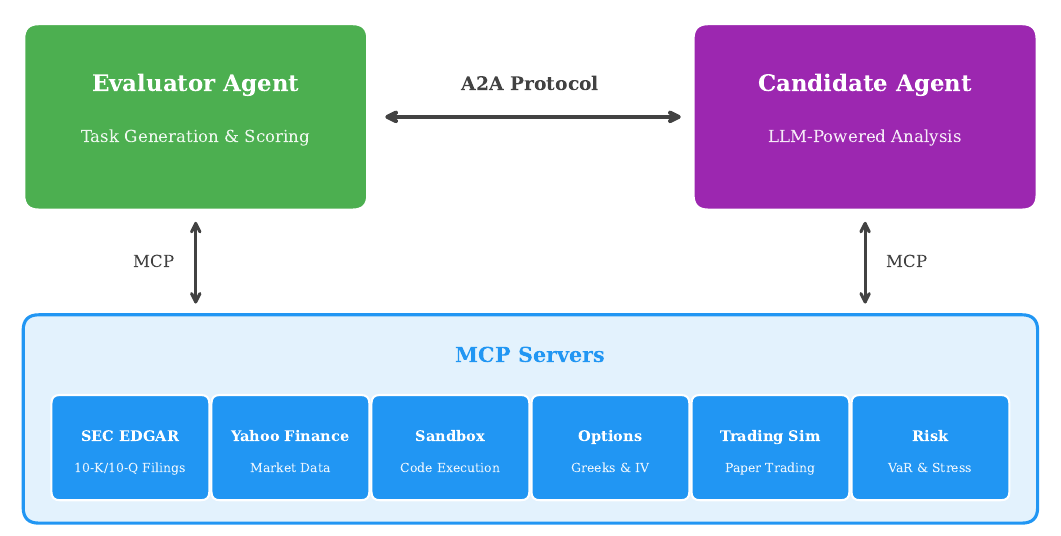}
\caption{Overview of the \bench{} two-agent architecture. The Evaluator Agent generates tasks from six datasets and sends them to the Candidate Agent via the A2A protocol. The Candidate Agent uses an LLM and six MCP servers to access financial data, execute code, and simulate trades. Responses are scored by dataset-specific evaluators.}
\label{fig:architecture}
\end{figure}

\bench{} implements a two-agent evaluation architecture (Figure~\ref{fig:architecture}) built on the AgentBeats platform \citep{agentbeats2025}. The Evaluator Agent orchestrates the benchmark: it loads evaluation configurations, generates tasks from six datasets, sends them to the candidate agent via the A2A protocol \citep{a2a2025}, and scores responses using dataset-specific evaluators. The Candidate Agent is the system under test---it receives financial analysis tasks and must produce responses using an underlying LLM and, critically, six specialized MCP servers \citep{mcp2024} that provide financial data access.

\begin{table}[H]
\centering
\caption{The six MCP servers available to the Candidate Agent.}
\label{tab:mcp}
\small
\begin{tabular}{lll}
\toprule
\textbf{Server} & \textbf{Function} & \textbf{Key Feature} \\
\midrule
SEC EDGAR & 10-K/10-Q filings & Temporal locking (no lookahead) \\
Yahoo Finance & Market data & Lookahead detection \\
Sandbox & Python execution & Sandboxed computation \\
Options & Black-Scholes pricing & Greeks \& IV calculation \\
Trading Sim & Paper trading & Slippage modeling \\
Risk & Portfolio risk metrics & VaR, Sharpe/Sortino \\
\bottomrule
\end{tabular}
\end{table}

\subsection{Evaluation Sections}
\label{sec:sections}

Tasks are organized into four equally weighted sections (Table~\ref{tab:sections}), each targeting distinct capabilities. The equal 25\% weighting prevents agents from achieving high overall scores by excelling in only one area. The current evaluation uses ${\sim}$50 tasks sampled from a larger pool; future evaluations can scale to hundreds by enabling additional tasks per section.

\begin{table}[h]
\centering
\caption{Overview of the four \bench{} evaluation sections (25\% each). Tool Use indicates whether MCP server interaction is required to answer correctly.}
\label{tab:sections}
\small
\begin{tabular}{llcll}
\toprule
\textbf{Section} & \textbf{Datasets} & \textbf{Tasks} & \textbf{Evaluator} & \textbf{Tool Use} \\
\midrule
Knowledge Retrieval & BizFinBench, PRBench & 18 & Exact match + LLM rubric & Required \\
Analytical Reasoning & Synthetic (CFA-level) & 9 & LLM rubric (3 components) & Not required \\
Options Trading & Options Alpha & 9 & 4-dim scoring & Required \\
Crypto Trading & Crypto (4 transforms) & 6 & Performance-based & Minimal \\
\bottomrule
\end{tabular}
\end{table}

Knowledge Retrieval tests whether agents can accurately extract financial facts and perform quantitative computations using real company data. BizFinBench \citep{bizfinbench2025} provides bilingual event logic reasoning and financial computation tasks, while PRBench \citep{prbench2025} tests multi-step professional reasoning. Both require agents to retrieve data from SEC filings and market data servers---agents answering from parametric knowledge alone produce stale or incorrect figures.

Analytical Reasoning evaluates self-contained financial computation through multi-step problems covering NPV and discount rate analysis, portfolio beta adjustment, bond pricing, free cash flow valuation, options strategies (put-call parity, spreads), leverage effects (Modigliani-Miller), binomial option pricing, duration immunization, and interest rate swaps. All information is provided in the question; no external data retrieval is needed, isolating reasoning from tool use. Scoring uses an LLM rubric with three components: methodology (30\%), calculation (30\%), and final answer (40\%).

Options Trading assesses quantitative derivatives knowledge through four sub-dimensions scored equally (25\% each): P\&L accuracy (max profit/loss, breakeven calculations), Greeks accuracy (delta, gamma, theta, vega within 5\% tolerance), strategy quality (multi-leg construction and rationale), and risk management (position sizing and hedging).

Crypto Trading evaluates agents under adversarial market conditions (Section~\ref{sec:adversarial}). Performance is measured by a weighted combination of total return (35\%), Sharpe ratio (30\%), win rate (20\%), and maximum drawdown penalty (15\%), aggregated across four progressively adversarial transform conditions.

\subsection{Adversarial Crypto Trading}
\label{sec:adversarial}

The crypto trading section applies four progressive data transforms to historical cryptocurrency price data, testing whether agents adapt their trading strategies to deteriorating market conditions:

\begin{table}[t]
\centering
\caption{Adversarial transform conditions applied to historical crypto price data. Each transform progressively increases the difficulty of signal extraction.}
\label{tab:transforms}
\small
\begin{tabular}{lp{9cm}}
\toprule
\textbf{Transform} & \textbf{Description} \\
\midrule
Baseline & Clean historical price data. Control condition for measuring base performance. \\
Noisy & Gaussian price noise ($\sigma = 2\%$) and sporadic volume spikes ($3\times$ normal). Tests microstructure robustness. \\
Meta & Combined noise patterns with trend modifications, false breakouts, and support/resistance violations. \\
Adversarial & Coordinated false signals targeting common strategies: moving average crossovers, RSI divergences, and MACD signal injections. \\
\bottomrule
\end{tabular}
\end{table}

Transform scores are weighted to emphasize baseline performance while penalizing adversarial fragility: baseline (40\%), noisy (30\%), adversarial (20\%), meta (10\%). This weighting reflects a deployment-oriented priority: an agent that performs well under normal conditions and degrades gracefully under attack is preferable to one that is uniformly mediocre.

\subsection{Unified Scoring}

All evaluator outputs are normalized to a $[0, 100]$ scale. For each section $s$, the section score is the mean of normalized task scores: $S_s = \frac{1}{|T_s|} \sum_{t \in T_s} \text{norm}(\text{score}_t)$. The overall score is the weighted sum across active sections:
\begin{equation}
\text{Overall} = \sum_{s \in \mathcal{S}} w_s \cdot S_s, \quad w_s = 0.25 \;\forall s
\end{equation}
If any section has no tasks (e.g., in a focused evaluation), weights are redistributed proportionally among active sections.

\section{Experimental Setup}

\subsection{Models}

We evaluate 12 models spanning frontier proprietary systems, open-weight models, and fully open-source models (Table~\ref{tab:models}). All models use the same Candidate Agent infrastructure with identical MCP server access, ensuring differences reflect model capability rather than infrastructure variation. We additionally test two ablation variants: GPT-5.2 with web search augmentation and Qwen3-32B with extended thinking mode (discussed in Section~\ref{sec:tool_use}).

\begin{table}[t]
\centering
\caption{Models evaluated. All access the same MCP server infrastructure through the Candidate Agent. Web search and thinking ablations are discussed separately.}
\label{tab:models}
\small
\begin{tabular}{llll}
\toprule
\textbf{Model} & \textbf{Type} & \textbf{Size} & \textbf{Provider} \\
\midrule
\multicolumn{4}{l}{\textit{Proprietary}} \\
Gemini-3-Pro & Proprietary & -- & Google \\
Kimi-K2.5 & Proprietary & -- & Moonshot AI \\
GPT-5.2 & Proprietary & -- & OpenAI \\
GPT-4o & Proprietary & -- & OpenAI \\
Grok 4.1 Fast & Proprietary & -- & xAI \\
\midrule
\multicolumn{4}{l}{\textit{Open-weight}} \\
GPT-OSS-120B & Open-weight & 120B & OpenAI \\
GPT-OSS-20B & Open-weight & 20B & OpenAI \\
\midrule
\multicolumn{4}{l}{\textit{Open-source}} \\
Qwen3-32B & Open-source & 32B & Alibaba \\
Qwen3-30B-A3B & Open-source (MoE) & 30B (3B active) & Alibaba \\
Qwen3-8B & Open-source & 8B & Alibaba \\
Gemma3-27B & Open-source & 27B & Google \\
GLM-4.7-Flash & Open-source (MoE) & 30B (3B active) & Zhipu AI \\
\bottomrule
\end{tabular}
\end{table}

\subsection{Evaluation Configuration}

All evaluations use the same configuration: approximately 50 tasks drawn by stratified sampling (seed 42), of which 42 are scored across the four main sections; the remaining professional tasks are analyzed separately (Appendix~\ref{app:professional}). The LLM judge for rubric-based evaluation sections is GPT-5.2 (temperature 0.0) for main results. The judge reliability study (Section~\ref{sec:judge}) re-evaluates identical responses with five additional judge models. Evaluations run with a 6-hour timeout per model.

\section{Results and Analysis}

\subsection{Overall Performance and Capability Stratification}

\begin{table}[t]
\centering
\caption{Main results. Models sorted by overall score (mean of four sections at 25\% each). \textbf{Bold}: best in column. \underline{Underline}: second best. KR = Knowledge Retrieval, AR = Analytical Reasoning, Opt = Options Trading, Cry = Crypto Trading.}
\label{tab:leaderboard}
\small
\begin{tabular}{lccccc}
\toprule
\textbf{Model} & \textbf{Overall} & \textbf{KR} & \textbf{AR} & \textbf{Opt} & \textbf{Cry} \\
\midrule
Gemini-3-Pro & \textbf{64.3} & 52.4 & \textbf{94.8} & 63.2 & 46.6 \\
Grok 4.1 Fast & \underline{63.7} & \textbf{61.6} & 87.9 & \textbf{72.2} & 33.2 \\
GPT-5.2 & 61.9 & \underline{50.6} & 87.3 & 62.1 & \underline{47.4} \\
Kimi-K2.5 & 54.4 & 37.1 & 71.0 & 62.6 & 46.8 \\
GLM-4.7-Flash & 53.9 & 38.9 & 82.1 & 61.0 & 33.5 \\
GPT-4o & 53.3 & 32.3 & 80.4 & 55.6 & 44.9 \\
GPT-OSS-120B & 50.9 & 32.5 & 74.8 & \underline{63.2} & 33.1 \\
GPT-OSS-20B & 50.6 & 42.4 & 72.1 & 55.4 & 32.6 \\
Qwen3-32B & 48.4 & 14.4 & 84.0 & 62.3 & 32.8 \\
Gemma3-27B & 47.2 & 9.3 & 68.7 & 59.0 & \textbf{51.7} \\
Qwen3-8B & 44.9 & 10.5 & 77.0 & 58.9 & 33.1 \\
Qwen3-30B-A3B & 44.7 & 9.4 & 74.1 & 62.6 & 32.7 \\
\bottomrule
\end{tabular}
\end{table}

\begin{figure}[t]
\centering
\includegraphics[width=0.90\textwidth]{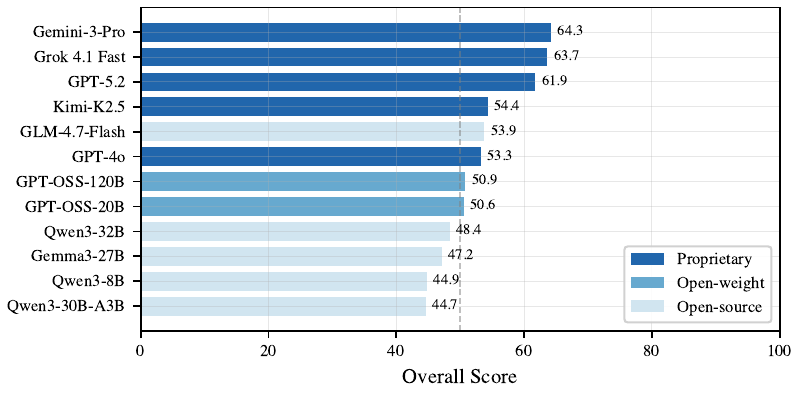}
\caption{Overall \bench{} scores for all 12 models, sorted by performance. The dashed line marks the 50/100 midpoint. Frontier proprietary models (left) clearly separate from smaller open-source models (right).}
\label{fig:overall}
\end{figure}

Table~\ref{tab:leaderboard} and Figure~\ref{fig:overall} present the full leaderboard. Two findings emerge. First, there is a clear tier structure: Gemini-3-Pro leads at 64.3 driven by the highest Analytical Reasoning (94.8) and strong Crypto (46.6), closely followed by Grok 4.1 Fast (63.7) with the best Knowledge Retrieval (61.6) and Options (72.2), then GPT-5.2 at 61.9, mid-tier proprietary and open-weight models at 50--54, and smaller open-source models at 44--48. The 20-point gap between best and worst demonstrates substantial capability stratification.

Second, the section that drives this gap is Knowledge Retrieval (range: 9.3--61.6). In contrast, Analytical Reasoning shows far less variance (68.7--94.8), and Options Trading is remarkably consistent (55.4--72.2). This reveals that the differentiating factor between models is not raw reasoning ability, but effective tool use for data retrieval.

\subsection{Knowledge Retrieval: Tool Use as the Key Differentiator}
\label{sec:tool_use}

Knowledge Retrieval has by far the highest cross-model variance (std.\ dev.\ 17.6 vs.\ 8.0 for AR), since it requires retrieving financial data from SEC filings and market data servers via MCP tools---agents answering from parametric knowledge alone produce stale or incorrect figures.

\begin{figure}[t]
\centering
\includegraphics[width=0.90\textwidth]{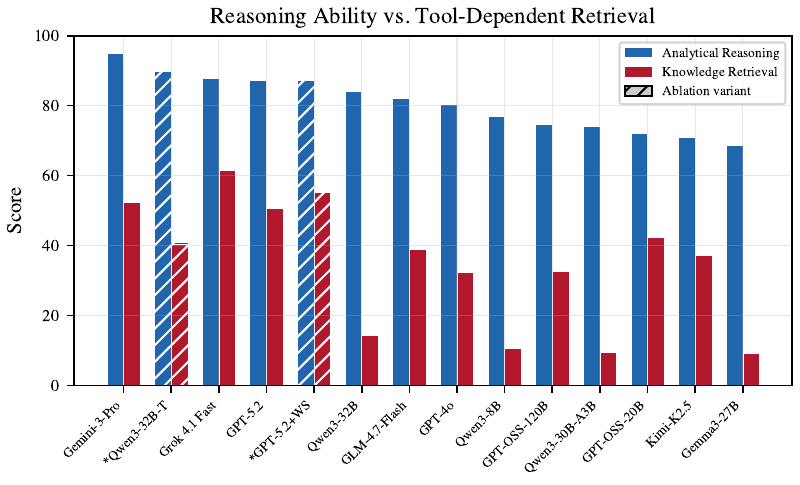}
\caption{Analytical Reasoning (self-contained) vs.\ Knowledge Retrieval (tool-dependent) scores. Models sorted by AR; hatched bars denote ablation variants (GPT-5.2+WS = web search, Qwen3-32B-T = thinking mode). Several base models score 75+ on reasoning but below 15 on retrieval. Both ablation variants show large KR gains, confirming that tool access and tool-use planning drive retrieval performance.}
\label{fig:kr_ar}
\end{figure}

Figure~\ref{fig:kr_ar} illustrates a stark divide: models that effectively use MCP tools (GPT-5.2: 50.6, Gemini-3-Pro: 52.4) dramatically outperform those that do not (Qwen3-8B: 10.5, Gemma3-27B: 9.3). Critically, models scoring below 15 on KR are not weak reasoners---Qwen3-32B scores 84.0 on AR (4th overall) but only 14.4 on KR (9th overall), demonstrating that model capability is necessary but not sufficient.

Two ablations confirm this. Adding a web search MCP server to GPT-5.2 yields $+$4.6 on KR (50.6 $\to$ 55.2). Enabling chain-of-thought in Qwen3-32B produces $+$26.1 on KR (14.4 $\to$ 40.5), the largest single-section improvement in our study, suggesting that extended reasoning improves multi-step tool-use planning. For tool-dependent tasks, improving tool access and planning matters more than model scale.

\subsection{The Conceptual-vs-Computational Gap in Options Trading}

\begin{figure}[t]
\centering
\includegraphics[width=0.90\textwidth]{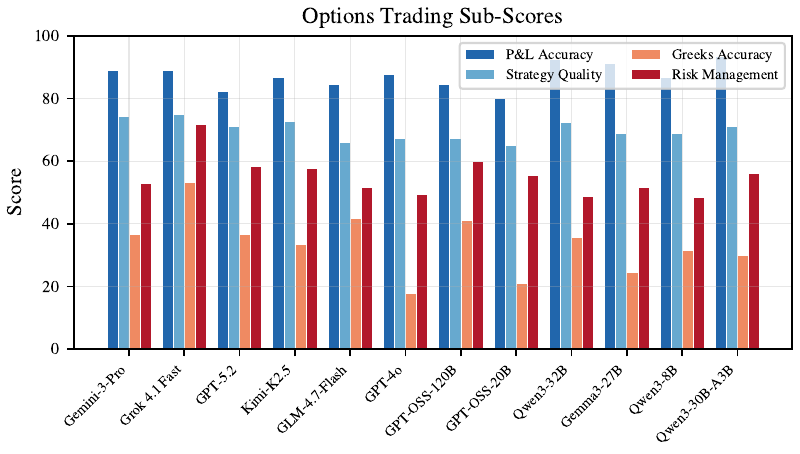}
\caption{Options trading sub-scores across all 12 models. P\&L accuracy (80--93) consistently dominates Greeks precision (18--53), revealing a universal conceptual-vs-computational gap with a mean 54-point difference. Note the "competence mirage" where models correctly identify strategies but fail to quantify their risks.}
\label{fig:options}
\end{figure}

Options Trading overall scores range 55.4--72.2, but sub-dimensions reveal a striking and systemic pattern (Figure~\ref{fig:options}). Across all 12 models, performance on conceptual tasks such as P\&L accuracy (80--93) and Strategy quality (65--75) consistently and significantly exceeds performance on computational tasks like Greeks precision (18--53) and Risk management (48--72).

This conceptual-vs-computational gap is pervasive and highlights a fundamental cognitive dissonance in current LLMs: models operate effectively as "semantic strategists" but fail as "numerical analysts." 
Specifically, models can correctly identify complex setups (e.g., identifying that an Iron Condor is appropriate for low-volatility environments) and calculate arithmetic expiration payoffs. However, they struggle profoundly to compute derivative sensitivities (delta, gamma, theta, vega), which require understanding instantaneous rates of change. 

Crucially, this failure persists even when models are equipped with an options pricing MCP server. This suggests the bottleneck is not merely internal computational capacity, but an \textit{interface failure}: models struggle to correctly parameterize external calls (e.g., aligning expiration dates, correctly estimating implied volatility inputs) or parse the high-precision numerical outputs returned by tools.

The disparity is most acute in state-of-the-art models. GPT-4o exhibits the largest gap (P\&L 87.8 vs.\ Greeks 17.8), creating a dangerous "competence mirage" where high-level reasoning masks low-level calculation failures. While Grok 4.1 Fast partially closes this gap with the highest Greeks score (53.3), a substantial 35.6-point difference remains. The safety implication is direct and critical: an agent that constructs a theoretically "hedged" strategy (e.g., a Delta-neutral portfolio) but computes Greeks incorrectly will inadvertently expose the portfolio to significant directional risk, all while confidently asserting the position is safe.

\subsection{Adversarial Robustness in Crypto Trading}

\begin{figure}[t]
\centering
\includegraphics[width=0.90\textwidth]{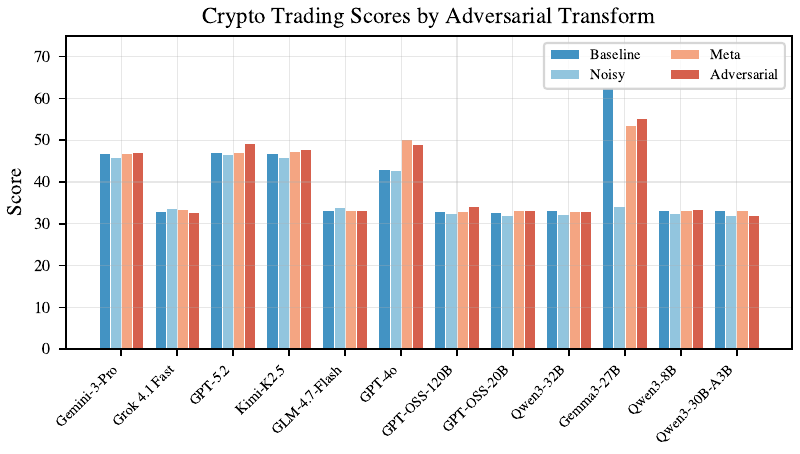}
\caption{Crypto trading scores by adversarial transform condition. Seven models (\emph{bottom cluster}, $\sim$32--34) show virtually no variation, suggesting a fixed non-adaptive strategy. Five models (\emph{top cluster}, 45--52) actively trade; among these, only GPT-4o and Gemma3-27B show large transform-dependent variation.}
\label{fig:crypto}
\end{figure}

The crypto trading section reveals a binary pattern across models (Figure~\ref{fig:crypto}). Seven of twelve models score between 32 and 34 across \emph{all four} transforms---including Grok 4.1 Fast (33.2), the second-ranked model overall---with negligible variation ($<$1 point between baseline and adversarial conditions). This flat profile suggests these models adopt a fixed strategy---likely minimal trading or buy-and-hold---that is trivially ``robust'' by being inert.

In contrast, five models form a \emph{top cluster} that actively trades: GPT-5.2 ($\sim$47), Gemini-3-Pro ($\sim$47), Kimi-K2.5 ($\sim$47), GPT-4o ($\sim$45), and Gemma3-27B ($\sim$52). Three of these---GPT-5.2, Gemini-3-Pro, and Kimi-K2.5---maintain consistently elevated scores across all transforms (range $<$2.5 points), suggesting a stable active strategy unaffected by signal manipulation. The remaining two show striking signal dependence: Gemma3-27B exhibits the widest spread (62.7 baseline, 34.2 noisy---a 28-point drop), while GPT-4o scores \emph{higher} under adversarial (49.1) and meta (50.1) than baseline (43.1), possibly reflecting contrarian positioning.

Gemma3-27B is particularly notable: despite ranking 10th overall (47.2), it achieves the \emph{highest} crypto score (51.7), driven by a baseline that far exceeds other models. However, its 28-point noisy-condition collapse reveals that the same signal sensitivity enabling high baseline performance makes it the most exploitable model under adversarial conditions.

This distinction between robustness through inaction and genuine adversarial resilience is critical for deployment. No model achieves both high baseline performance and minimal adversarial degradation---resolving this tension remains an open challenge.

\subsection{Judge Reliability: When Evaluation Itself is Unreliable}
\label{sec:judge}

\begin{table}[H]
\centering
\caption{Judge comparison: identical GPT-5.2 Candidate Agent responses evaluated by three different judge models. Even with only three judges, Knowledge Retrieval scores vary by nearly 29 points.}
\label{tab:judge}
\small
\begin{tabular}{lccccc}
\toprule
\textbf{Judge Model} & \textbf{Overall} & \textbf{KR} & \textbf{AR} & \textbf{Opt} & \textbf{Cry} \\
\midrule
Gemini-3-Flash & 66.5 & 78.2 & 81.7 & 58.8 & 47.1 \\
GPT-5.2 (baseline) & 61.9 & 50.6 & 87.3 & 62.1 & 47.4 \\
Claude Sonnet 4.5 & 55.2 & 49.4 & 68.3 & 55.7 & 47.4 \\
\midrule
\textit{Range} & \textit{11.3} & \textit{28.8} & \textit{19.0} & \textit{6.4} & \textit{0.3} \\
\bottomrule
\end{tabular}
\end{table}

To assess evaluation reliability, we held the Candidate Agent's responses constant (GPT-5.2) and re-evaluated them with three different judge models (Table~\ref{tab:judge}). Even with only three frontier-class judges, significant judge-dependent variance emerges.

\begin{figure}[t]
\centering
\includegraphics[width=0.80\textwidth]{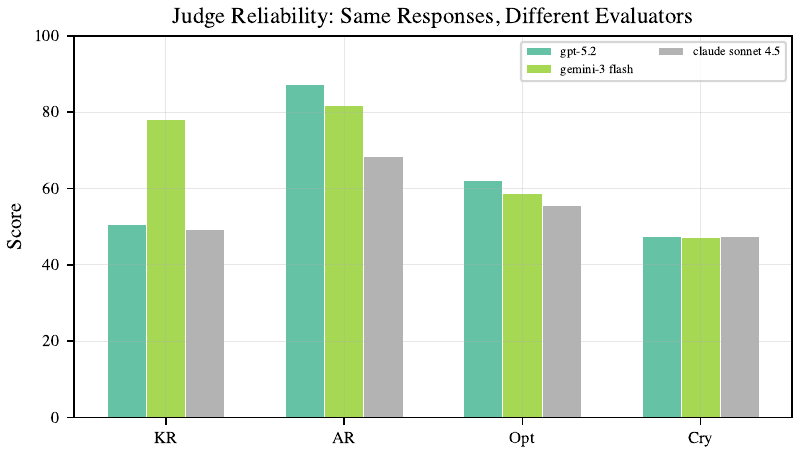}
\caption{Per-section score distributions across three judge models evaluating identical responses. Crypto Trading (performance-based scoring) shows near-zero variance, while Knowledge Retrieval (LLM rubric scoring) shows substantial judge disagreement.}
\label{fig:judge}
\end{figure}

Crypto is the most judge-invariant section (range: 0.3 points), because it uses performance-based metrics rather than LLM judgment. Knowledge Retrieval is the most variable (range: 28.8 points)---Gemini-3-Flash assigns 78.2 while Claude Sonnet 4.5 assigns 49.4 to the \emph{same responses}. Options Trading (range: 6.4) achieves the best agreement among LLM-judged sections, likely because verifiable numerical components constrain judge discretion. Figure~\ref{fig:judge} visualizes these patterns.

The safety implication is direct: if we cannot reliably evaluate agents, we cannot reliably deploy them. For safety-critical financial applications, this argues for performance-based metrics over LLM judgment where possible, multi-judge evaluation protocols, and explicit judge calibration against human expert assessments.

\section{Discussion}

\paragraph{Safety implications.}
Three findings connect directly to the safe deployment of finance agents. First, the adversarial crypto results reveal that most models achieve apparent robustness through \emph{inaction}, not genuine resilience---a distinction invisible to aggregate score comparisons. Deploying such agents could create a false sense of security. Second, the conceptual-vs-computational gap in options trading means agents may \emph{correctly describe} a hedging strategy while \emph{incorrectly computing} its parameters---a failure mode that produces plausible-sounding but quantitatively wrong risk assessments. Third, the 11-point judge spread across three frontier models demonstrates that evaluation reliability is itself a safety concern: deployment decisions based on single-judge benchmarks are unreliable.

\paragraph{Limitations and future work.}
The current evaluation uses ${\sim}$50 tasks per model, sampled from a larger pool that can scale to hundreds in future iterations. Each model is evaluated in a single run; while judge temperature is set to 0 to minimize scoring variance, the Candidate Agent's own generation introduces stochasticity, and we do not yet report confidence intervals across multiple seeds.  Regarding data provenance, crypto trading scenarios are sampled from an extensive repository ($>$100GB) of real historical market data, rather than being synthetically generated. This vast temporal breadth allows the benchmark to be continuously refreshed with novel market intervals, effectively mitigating data contamination and overfitting. The Candidate Agent architecture is fixed across models, isolating LLM capability differences but not capturing potential gains from model-specific agent design. Future work will explore multi-seed evaluation with variance estimates and model-adaptive agent architectures.

\section{Conclusion}

We presented \bench{}, a hybrid benchmark designed to bridge the gap between static financial knowledge and dynamic market execution. By combining expert-verified static tasks with performance-grounded trading simulations, we eliminate the reliance on high-variance LLM judges for decision-making evaluation. Our empirical study of 13 models (ranging from 8B open-source to frontier reasoning models) reveals a critical disconnect in current AI capabilities:

\begin{enumerate}
    \item While "extended thinking" capabilities dramatically enhance knowledge retrieval (+26 points), they yield negligible improvements in dynamic execution (+0.3 in crypto, $-$0.1 in options). This suggests that current chain-of-thought paradigms improve information synthesis but fail to translate into better real-time market adaptation.
    \item In adversarial crypto scenarios, 8 of 13 models maintained static scores (${\sim}$33) with less than 1-point variation across four progressive market manipulations. This indicates that agents are not resiliently adapting to market shifts but are instead defaulting to fixed, non-adaptive strategies.
    \item The "conceptual-vs-computational gap" in options trading remains a persistent safety risk. By scoring agents on realized metrics (Sharpe ratio, Greeks precision) rather than semantic plausibility, \bench{} exposes failures that purely text-based evaluations miss.
\end{enumerate}

These findings argue that the path to autonomous finance agents lies not merely in scaling inference compute, but in fundamentally improving dynamic decision-making architectures. The \bench{} framework, dataset, and evaluation protocols are publicly available to facilitate this next step in agentic finance research.\footnote{Repository URL withheld for double-blind review.}

\bibliographystyle{iclr2026_conference}
\bibliography{iclr2026_conference}

\newpage
\appendix

\section{Options Trading Sub-Score Details}
\label{app:options}

\begin{table}[h]
\centering
\caption{Full options trading sub-scores for all models.}
\small
\begin{tabular}{lcccc}
\toprule
\textbf{Model} & \textbf{P\&L} & \textbf{Strategy} & \textbf{Greeks} & \textbf{Risk} \\
\midrule
Gemini-3-Pro & 88.9 & 74.4 & 36.7 & 52.8 \\
Grok 4.1 Fast & 88.9 & 75.0 & 53.3 & 71.7 \\
GPT-5.2 & 82.2 & 71.1 & 36.7 & 58.3 \\
Kimi-K2.5 & 86.7 & 72.8 & 33.3 & 57.8 \\
GLM-4.7-Flash & 84.4 & 66.1 & 41.7 & 51.7 \\
GPT-4o & 87.8 & 67.2 & 17.8 & 49.4 \\
GPT-OSS-120B & 84.4 & 67.2 & 41.1 & 60.0 \\
GPT-OSS-20B & 80.0 & 65.0 & 21.1 & 55.6 \\
Qwen3-32B & 92.5 & 72.5 & 35.6 & 48.8 \\
Gemma3-27B & 91.1 & 68.9 & 24.4 & 51.7 \\
Qwen3-8B & 86.7 & 68.9 & 31.7 & 48.3 \\
Qwen3-30B-A3B & 93.3 & 71.1 & 30.0 & 56.1 \\
\midrule
\multicolumn{5}{l}{\textit{Ablation variants}} \\
Qwen3-32B + Think & 91.1 & 70.6 & 32.2 & 55.0 \\
\bottomrule
\end{tabular}
\end{table}

\section{Crypto Trading Transform Details}
\label{app:crypto}

\begin{table}[h]
\centering
\caption{Full crypto trading scores by transform condition for all models.}
\small
\begin{tabular}{lcccc}
\toprule
\textbf{Model} & \textbf{Baseline} & \textbf{Noisy} & \textbf{Meta} & \textbf{Adversarial} \\
\midrule
Gemini-3-Pro & 46.9 & 45.8 & 46.9 & 47.2 \\
Grok 4.1 Fast & 32.9 & 33.8 & 33.5 & 32.7 \\
GPT-5.2 & 47.2 & 46.7 & 47.0 & 49.2 \\
Kimi-K2.5 & 46.8 & 46.0 & 47.3 & 47.9 \\
GLM-4.7-Flash & 33.3 & 33.9 & 33.1 & 33.2 \\
GPT-4o & 43.1 & 42.7 & 50.1 & 49.1 \\
GPT-OSS-120B & 32.9 & 32.6 & 33.0 & 34.3 \\
GPT-OSS-20B & 32.7 & 32.0 & 33.1 & 33.2 \\
Qwen3-32B & 33.2 & 32.2 & 33.1 & 33.0 \\
Gemma3-27B & 62.7 & 34.2 & 53.6 & 55.2 \\
Qwen3-8B & 33.3 & 32.6 & 33.2 & 33.5 \\
Qwen3-30B-A3B & 33.3 & 32.0 & 33.3 & 32.1 \\
\midrule
\multicolumn{5}{l}{\textit{Ablation variant}} \\
Qwen3-32B + Think & 33.1 & 32.3 & 33.2 & 34.1 \\
\bottomrule
\end{tabular}
\end{table}

\section{Professional Tasks: Correlation with Trading}
\label{app:professional}

We additionally evaluate all models on GDPVal \citep{gdpval2026}, a professional task benchmark spanning 44 occupations, to test whether general professional competence predicts trading ability. Table~\ref{tab:prof_corr} reports each model's Professional Tasks score alongside its trading composite (mean of Options and Crypto).

\begin{table}[h]
\centering
\caption{Professional Tasks (GDPVal) scores and trading composite. Pearson correlations: Prof vs.\ Trading $r{=}0.36$, Prof vs.\ Crypto $r{=}0.16$, Prof vs.\ KR $r{=}0.62$.}
\label{tab:prof_corr}
\small
\begin{tabular}{lccc}
\toprule
\textbf{Model} & \textbf{Prof} & \textbf{Trading} & \textbf{$\Delta$} \\
\midrule
Gemini-3-Pro & 29.2 & 54.9 & $-$25.7 \\
Grok 4.1 Fast & 84.8 & 52.7 & $+$32.1 \\
GPT-5.2 & 48.9 & 54.8 & $-$5.9 \\
Kimi-K2.5 & 30.7 & 54.7 & $-$24.0 \\
GLM-4.7-Flash & 16.2 & 47.2 & $-$31.0 \\
GPT-4o & 63.1 & 50.2 & $+$12.9 \\
GPT-OSS-120B & 29.8 & 48.2 & $-$18.4 \\
GPT-OSS-20B & 36.0 & 44.0 & $-$8.0 \\
Qwen3-32B & 15.3 & 47.5 & $-$32.2 \\
Gemma3-27B & 20.3 & 55.4 & $-$35.1 \\
Qwen3-8B & 24.5 & 46.0 & $-$21.5 \\
Qwen3-30B-A3B & 29.8 & 47.7 & $-$17.9 \\
\midrule
\multicolumn{4}{l}{\textit{Ablation variants}} \\
GPT-5.2 + WS & 51.2 & 53.7 & $-$2.5 \\
Qwen3-32B + Think & 28.0 & 47.6 & $-$19.6 \\
\bottomrule
\end{tabular}
\end{table}

Professional Tasks correlates moderately with Knowledge Retrieval ($r{=}0.62$), as both reward tool-use and output structuring. However, the correlation with trading is weak ($r{=}0.36$) and nearly absent for Crypto ($r{=}0.16$). Notable outliers include Grok~4.1~Fast (Prof~84.8, Crypto~33.2) and Gemma3-27B (Prof~20.3, Crypto~51.7), confirming that professional output quality and adversarial trading resilience measure fundamentally different capabilities. This supports our decision to focus the \bench{} scoring on four trading-oriented sections.

\section{Synthetic Question Topics}
\label{app:questions}

The Analytical Reasoning section draws from 22 financial computation questions spanning 10 topic areas: Capital Budgeting (2), Portfolio Theory (3), Fixed Income (4), Corporate Finance (3), Options \& Derivatives (4), Time Value of Money (2), Valuation (2), Forex (1), Corporate Actions (1), and Combined Leverage (1). Questions are self-contained with all necessary information provided and have unambiguous correct answers.

\end{document}